\pdfoutput=1
\documentclass[sigconf]{acmart}

\ifdefined\pdftexversion\else
  \makeatletter
  \@ifundefined{author@bx@sep}{\newskip\author@bx@sep}{}
  \@ifundefined{author@bx@wd}{\newdimen\author@bx@wd}{}
  \@ifundefined{@mkauthors@iii}{\def\@mkauthors@iii{}}{}
  \@ifundefined{@typeset@author@bx}{\def\@typeset@author@bx{}}{}
  \makeatother
  \providecommand{\setcctype}[2][]{}
\fi

\makeatletter
\patchcmd{\@mkauthors@iii}
  {\mathchardef\UrlBreakPenalty=10000}
  {\mathchardef\UrlBreakPenalty=1000}{}{}
\patchcmd{\@mkauthors@iii}
  {\mathchardef\UrlBreakPenalty=10000}
  {\mathchardef\UrlBreakPenalty=1000}{}{}

\def\UrlBigBreaks{\do\:\do@url@hyp\do\@}
\mathchardef\UrlBigBreakPenalty=100
\newcount\acm@authorbreak@count
\newdimen\acm@narrow@wd
\let\acm@orig@typeset@author@bx\@typeset@author@bx
\def\@typeset@author@bx{%
  \ifnum\acm@authorbreak@count=0\relax
    \acm@narrow@wd=\author@bx@wd
    \author@bx@wd=0.32\textwidth\relax
  \fi
  \acm@orig@typeset@author@bx
  \global\advance\acm@authorbreak@count by 1\relax
  \ifnum\acm@authorbreak@count=3\relax
    \author@bx@wd=\acm@narrow@wd
    \break
  \fi
}
\makeatother

\AtBeginDocument{}
\providecommand{\setcctype}[2][]{}

\acmYear{2026}\copyrightyear{2026}
\setcopyright{cc}
\setcctype[4.0]{by}
\acmConference[ACM CAIS '26]{ACM Conference on AI and Agentic Systems}{May 26--29, 2026}{San Jose, CA, USA}
\acmBooktitle{ACM Conference on AI and Agentic Systems (ACM CAIS '26), May 26--29, 2026, San Jose, CA, USA}
\acmDOI{10.1145/3786335.3813164}
\acmISBN{979-8-4007-2415-2/26/05}

\newcommand{\paperreview}{review}
\newcommand{\paperpreprint}{preprint}
\newcommand{\paperfinal}{final}
\let\paperversion\paperfinal

\newcommand{\visibilityanonymous}{anonymous}
\newcommand{\visibilityrevealed}{revealed}

\ifx\paperversion\paperreview
    \let\papervisibility\visibilityanonymous
\else\ifx\paperversion\paperpreprint
    \let\papervisibility\visibilityrevealed
\else\ifx\paperversion\paperfinal
    \let\papervisibility\visibilityrevealed
\fi\fi\fi

\usepackage{fix-cm}
\usepackage[utf8]{inputenc}
\usepackage{latexsym}
\usepackage{amsmath,amsthm,mathtools}
\usepackage{nicefrac}

\usepackage{microtype}
\usepackage{xcolor}

\usepackage{hyperref}
\usepackage{xurl}

\usepackage{algorithm}
\usepackage[noend]{algpseudocode}
\usepackage{listingsutf8}
\usepackage{fontawesome5}

\usepackage{graphicx}
\usepackage{subfigure}
\usepackage{float}

\usepackage{booktabs}
\usepackage{multirow}
\usepackage{longtable}
\usepackage{array}
\usepackage[most]{tcolorbox}

\definecolor{CustomRed}{HTML}{E30A17}
\definecolor{DarkCyan}{HTML}{0E9594}
\definecolor{OrangeWeb}{HTML}{FFA400}
\definecolor{Avocado}{HTML}{688E26}
\definecolor{TyrianPurple}{HTML}{550527}
\definecolor{BurntUmber}{HTML}{772E25}
\definecolor{FireEngineRed}{HTML}{C33626}
\definecolor{Carmine}{HTML}{990011}
\definecolor{CaribbeanCurrent}{HTML}{187077}
\definecolor{LightGreen}{HTML}{D2f4d3}
\definecolor{PigmentGreen}{HTML}{08A045}
\definecolor{VanillaPink}{HTML}{E57a81}
\definecolor{EarthYellow}{HTML}{E0A458}

\definecolor{PastelBlue}{HTML}{1E70EB}
\definecolor{PastelGreen}{HTML}{238636}
\definecolor{PastelPurple}{HTML}{8957E5}
\definecolor{PastelYellow}{HTML}{9f6B01}
\definecolor{PastelRed}{HTML}{DA3532}

\definecolor{PastelOrange}{HTML}{F66A0A}
\definecolor{PastelPink}{HTML}{EA4AAA}
\definecolor{PastelTeal}{HTML}{1B7C83}
\definecolor{PastelCyan}{HTML}{00B4D8}
\definecolor{PastelIndigo}{HTML}{5A32A3}
\definecolor{PastelBrown}{HTML}{7F4E1E}
\definecolor{PastelOlive}{HTML}{6C7E00}
\definecolor{PastelGray}{HTML}{586069}
\definecolor{PastelTurquoise}{HTML}{30CFCF}
\definecolor{PastelCoral}{HTML}{FF6F61}

\colorlet{PrimaryColor}{CaribbeanCurrent}
\colorlet{SecondaryColor}{EarthYellow}

\colorlet{LinkColor}{PrimaryColor}
\colorlet{mmGRPOColor}{PrimaryColor!70}
\colorlet{HighlightColorQ}{PastelYellow!30}
\colorlet{HighlightColorTheta}{PastelTeal!30}
\colorlet{HighlightColorTeachers}{PastelYellow!30}
\colorlet{HighlightColorModuleLevel}{PastelTeal!30}

\newcounter{snippetpython}

\makeatletter
\lstnewenvironment{snippetpython}[1][]{
    \let\ClassWarning\@gobbletwo
    
    \let\c@lstlisting=\c@snippetpython
    
    \lstset{
        language=Python,
        basicstyle=\ttfamily\footnotesize,
        captionpos=b,
        showstringspaces=false,
        breaklines=true,
        numbers=left,
        numberstyle=\tiny\color{gray},
        keywordstyle=\color{DarkCyan},
        stringstyle=\color{OrangeWeb},
        commentstyle=\color{Avocado},
        morekeywords={dspy, self},
        backgroundcolor=\color{gray!5},
        numbersep=5pt,
        xleftmargin=20pt,
        xrightmargin=20pt,
        frame=single,
        framerule=0pt,
        framesep=10pt,
        framexleftmargin=10pt,
        framexrightmargin=10pt,
        #1
    }
}{}
\makeatother

\newcounter{datapoint}
\newsavebox{\datapointbox}
\newenvironment{datapoint}[1]{
    \refstepcounter{datapoint}
    \par\addvspace{\smallskipamount}
    \noindent
    \setlength{\fboxsep}{5pt}%
    \setlength{\fboxrule}{1pt}%
    \begin{lrbox}{\datapointbox}
    \begin{minipage}{\dimexpr\linewidth-2\fboxsep-2\fboxrule\relax}
        \ttfamily\footnotesize
        \textbf{#1}\par\addvspace{\smallskipamount}
    }
    {
    \end{minipage}%
    \end{lrbox}%
    \fcolorbox{SecondaryColor!70}{SecondaryColor!10}{\usebox{\datapointbox}}%
    \par\addvspace{\smallskipamount}
}

\newcounter{chatmessage}
\makeatletter
\lstnewenvironment{chatmessage}[1][]{%
    \let\ClassWarning\@gobbletwo
    
    \let\c@lstlisting=\c@chatmessage
    
    \lstset{
        basicstyle=\ttfamily\small,
        breaklines=true,
        breakindent=0pt,
        breakautoindent=false,
        columns=fullflexible,
        frame=single,
        framerule=0pt,
        framesep=10pt,
        framexleftmargin=10pt,
        framexrightmargin=10pt,
        backgroundcolor=\color{gray!5},
        captionpos=b,
        xleftmargin=20pt,
        xrightmargin=20pt,
        #1
    }
}{}
\makeatother
\newcommand{\autorefchatmessage}[1]{\hyperref[#1]{Message~\ref*{#1}}}

\newcommand{\WrappedAlgorithmCommand}[1]{%
  \parbox[t]{\dimexpr\linewidth - (\algorithmicindent * 3)\relax}{#1}%
}

\algnewcommand{\WrappedState}[1]{%
  \State \WrappedAlgorithmCommand{#1}%
}

\algnewcommand{\LeftComment}[1]{%
  \State \WrappedAlgorithmCommand{\(\triangleright\) #1}
}

\algnewcommand{\IdeaComment}[1]{%
  \State \WrappedAlgorithmCommand{\(\triangleright\) #1}
}

\hypersetup{
    colorlinks=true,
    allcolors={LinkColor},
    linkcolor={LinkColor},
    citecolor={LinkColor},
    breaklinks=true,
}

\newcommand{\mmgrpo}{\textcolor{mmGRPOColor}{\textsc{mmGRPO}}}

\newcommand{\bankingtext}{Banking77}
\newcommand{\papillontext}{PAPILLON}
\newcommand{\hovertext}{HoVer}
\newcommand{\banking}{\texttt{\bankingtext{}}}
\newcommand{\papillon}{\texttt{\papillontext{}}}
\newcommand{\hover}{\texttt{\hovertext{}}}
\newcommand{\hoverfour}{\texttt{\hovertext{}\textsubscript{4-HOP}}}

\newcommand{\colberttwo}{\texttt{ColBERTv2}}

\newcommand{\llamathreepointone}{\texttt{meta-llama/Meta-Llama-3.1-8B-Instruct}}
\newcommand{\llamathreepointoneshort}{\texttt{llama3.1-8b-instruct}}
\newcommand{\llamathreepointoneshorter}{\texttt{llama3.1}}

\newcommand{\qwenthree}{\texttt{Qwen/Qwen3-8B}}
\newcommand{\qwenthreeshort}{\texttt{qwen3-8b}}
\newcommand{\qwenthreeshorter}{\texttt{qwen3}}

\newcommand{\gptfouronemini}{\texttt{openai/gpt-4.1-mini-2025-04-14}}

\newcommand{\apachetwolicense}{\texttt{Apache License 2.0}}
\newcommand{\mitlicense}{\texttt{MIT License}}
\newcommand{\ccbyfour}{\texttt{CC BY 4.0}}
\newcommand{\llamathreecommunitylicense}{\texttt{Meta Llama 3 Community License}}

\newcommand{\gpuhonehundred}{\texttt{H100}}

\newcommand{\gainspovscot}{6\%}
\newcommand{\gainsmmgrpovscot}{7\%}

\newcommand{\gainsbettertogethervscot}{11\%}
\newcommand{\gainsbettertogethervspo}{5\%}
\newcommand{\gainsbettertogethervsmmgrpo}{3\%}

\begin{document}

\title{Composing Policy Gradients and Prompt Optimization for Language Model Programs}

\author{Noah Ziems}
\authornote{All three authors contributed equally to this research. Correspondence to \href{mailto:nziems2@nd.edu}{nziems2@nd.edu}, \href{mailto:soylu@stanford.edu}{soylu@stanford.edu}, and \href{mailto:lakshyaaagrawal@berkeley.edu}{lakshyaaagrawal@berkeley.edu}.}
\affiliation{%
  \institution{University of Notre~Dame}
  \city{Notre Dame}
  \country{USA}
}
\email{nziems2@nd.edu}

\author{Dilara Soylu}
\authornotemark[1]
\affiliation{%
  \institution{Stanford University}
  \city{Stanford}
  \country{USA}
}
\email{soylu@stanford.edu}

\author{Lakshya A Agrawal}
\authornotemark[1]
\affiliation{%
  \institution{UC Berkeley}
  \city{Berkeley}
  \country{USA}
}
\email{lakshyaaagrawal@berkeley.edu}

\author{Isaac Miller}
\affiliation{%
  \institution{Anyscale}
  \city{San Francisco}
  \country{USA}
}
\email{isaac@cmpnd.ai}

\author{Liheng Lai}
\affiliation{%
  \institution{UC Berkeley}
  \city{Berkeley}
  \country{USA}
}
\email{liheng@berkeley.edu}

\author{Chen Qian}
\affiliation{%
  \institution{CMU}
  \city{Pittsburgh}
  \country{USA}
}
\email{chenqian@cmu.edu}

\author{Kaiqiang Song}
\affiliation{%
  \institution{Zoom, Inc.}
  \city{San Jose}
  \country{USA}
}
\email{kaiqiang.song@zoom.us}

\author{Meng Jiang}
\affiliation{%
  \institution{\hspace*{-6pt}\mbox{University of Notre Dame}}
  \city{Notre Dame}
  \country{USA}
}
\email{mjiang2@nd.edu}

\author{Dan Klein}
\affiliation{%
  \institution{UC Berkeley}
  \city{Berkeley}
  \country{USA}
}
\email{klein@cs.berkeley.edu}

\author{Matei Zaharia}
\affiliation{%
  \institution{UC Berkeley}
  \city{Berkeley}
  \country{USA}
}
\email{matei@berkeley.edu}

\author{Karel D'Oosterlinck}
\affiliation{%
  \institution{Contextual AI}
  \city{Mountain View}
  \country{USA}
}
\email{karel@contextual.ai}

\author{Christopher Potts}
\affiliation{%
  \institution{Stanford University}
  \city{Stanford}
  \country{USA}
}
\email{cgpotts@stanford.edu}

\author{Omar Khattab}
\affiliation{%
  \institution{MIT}
  \city{Cambridge}
  \country{USA}
}
\email{okhattab@mit.edu}

\renewcommand{\shortauthors}{Ziems*, Soylu*, Agrawal* et al.}

\begin{abstract}
    Group Relative Policy Optimization (GRPO) has proven to be an effective tool for post-training language models (LMs).
However, AI systems are increasingly expressed as modular programs that mix together multiple LM calls with distinct prompt templates and other tools, and it is not clear how practitioners can best leverage online RL algorithms like GRPO to improve these systems.
We begin to address this challenge by investigating whether it is possible to effectively instantiate GRPO for arbitrary multi-prompt programs and whether it can work robustly as an off-the-shelf optimizer for LM programs using the same abstractions and constraints typically involved for prompt optimization.
Our main variant of multi-module GRPO constructs groups from module-level invocations, and we also consider trajectory-level grouping as another natural instantiation.
We find for the first time that GRPO (and its multi-module counterpart) empirically composes well with automatic prompt optimization, and together they improve accuracy by \gainsbettertogethervscot{} on average across classification, many-hop search, and privacy-preserving delegation tasks against the post-trained LM---with \gainsbettertogethervspo{} gains against prompt optimization on its own.
\ifx\papervisibility\visibilityanonymous
    Our approach is released as an open-source learning algorithm for compound AI systems.
\else
    We open-source multi-module GRPO in the DSPy library at \url{dspy.ai}.
\fi

\begin{center}
    \faGithub\,
    \ifx\papervisibility\visibilityanonymous  %
        \href{null}{\texttt{github.com/[redacted]/[redacted]}}
    \else
        \href{https://github.com/stanfordnlp/dspy}{\texttt{https://github.com/stanfordnlp/dspy}}
    \fi
\end{center}

\end{abstract}

\ccsdesc[500]{Computing methodologies~Natural language processing}
\ccsdesc[300]{Computing methodologies~Reinforcement learning}
\ccsdesc[100]{Computing methodologies~Multi-agent systems}

\keywords{Compound AI Systems, Language Program Optimization, Prompt Optimization, In-context Learning, Online Reinforcement Learning, BetterTogether}

\maketitle

\section{Introduction}
\label{sec:introduction}

Many modern AI systems and agents are increasingly implemented as modular designs, in which modules are responsible for well-specified subtasks that contribute to a broader objective.
A canonical example is ``multi-hop'' deep research, where the system responds to a question by iteratively using a \textit{query generation} LM module to produce a search query, passing that query to a retriever, and finally feeding all iteratively retrieved passages into a \textit{response generation} LM module to produce the final output~\citep{yang-etal-2018-hotpotqa, jiang-etal-2020-hover, nakano2021webgpt, qi2021iterative}.
Akin to conventional software, the explicit modularization of such systems makes their behavior controllable, testable, parallelizable, and allows for leveraging the priors of the LM differently for each module.
This modularization also allows for a separation of concerns: developers can iterate on their programs, and can bolt learning algorithms, typically automatic prompt optimization methods, on top to tune their quality~\citep{khattab2024dspy}.
Such optimizers can run the program, observe module-level inputs and outputs, score the final output with a program-level metric, and update the prompts or other parameters.

\begin{figure*}[tp]
    \centering
    \includegraphics[width=0.95\textwidth]{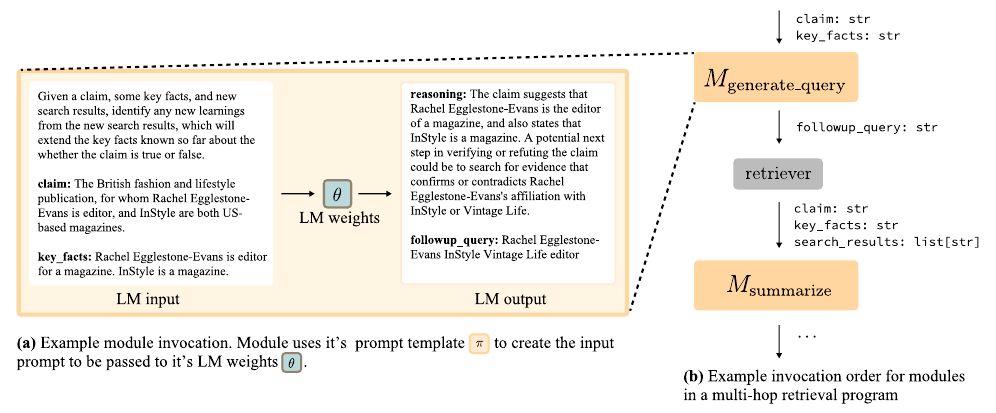}
    \caption{
        Simplified example of a multi-hop retrieval LM program invocation.
        Each module is responsible for a single subtask, and only its structured outputs (e.g., \texttt{followup\_query}) are passed to downstream modules.
        Bold text indicates the specific spans extracted as the module's output.
        Although each module internally generates tokens to produce these outputs, the raw generation traces (prompts, reasoning steps, and intermediate tokens) remain local to the module and are not shared across the pipeline.
        An example of a full trace can be found in \autoref{app:example_trace}.
    }
    \Description{A horizontal pipeline diagram showing a multi-hop retrieval LM program: a query-generation module produces a search query, a retriever returns passages, and a response-generation module produces the final answer. Each module is drawn as a box with its own local prompt and reasoning trace; only the structured outputs (e.g., a follow-up query string) are shown crossing between modules, while reasoning tokens stay inside each box. Bold text marks the specific output span passed downstream.}
    \label{fig:mm_rollout_diagram}
\end{figure*}

Group Relative Policy Optimization (GRPO; \citealt{shao2024deepseekmathpushinglimitsmathematical}) has recently emerged as a powerful method for fine-tuning language models (LMs) in the final stages of training.
By leveraging relative rewards within groups of ``reasoning'' rollouts that share the same prompt, GRPO offers a simple alternative to Proximal Policy Optimization (PPO; \citealt{schulman2017proximal}), with strong empirical results and far lower engineering overhead.
However, GRPO was originally designed for single-stage settings where each rollout consists of a single auto-regressive LM call.
It remains unclear how to best instantiate post-training RL algorithms such as GRPO for compound systems that involve multiple LM calls whose contexts, control flow, and intermediate states differ across rollouts.
As in standard GRPO, rewards are typically only observed at the level of full-trajectory rollouts.
However, as each module has its own isolated context, applying GRPO to the multi-module setting introduces challenges that do not arise in single-call systems or in simple LLM agents, like ReAct, where a single chat history captures the full trajectory.
For example, rollouts from the same LM program input can differ in both length and structure due to control-flow variation, early termination, or parsing failures.
An example of an LM program invocation can be seen in \autoref{fig:mm_rollout_diagram}.

To investigate whether GRPO can effectively be instantiated for such LM programs with a simple and extensible framework, we develop the \mmgrpo{} framework.
The main variant we test with \mmgrpo{} constructs module-level groups, relaxing the conventional GRPO implementation assumption of single-prompt groups by aligning structurally comparable module calls across different trajectories.
We also test a trajectory-level variant of \mmgrpo{} as another natural instantiation of GRPO for multi-module programs.
Both \mmgrpo{} variants can be used across a wide range of LM programs and are implemented to extend the existing abstractions for multi-module prompt optimization approaches, allowing them to be used interchangeably for improving program performance.
We open-source \mmgrpo{} as an off-the-shelf optimizer for arbitrary compound AI systems.

Because ours is the first implementation of GRPO that applies to sophisticated pipelines of LMs, we are able to conduct a \textit{controlled comparison} of three approaches to optimizing modular AI systems: prompt optimization (PO), online reinforcement learning via \mmgrpo{}, and their combination using the BetterTogether framework~\citep{soylu-etal-2024-fine}, which is our main contribution in this work.
Our evaluation setup comes from a subset of LangProBe, a Language Programs Benchmark \citep{tan2025langprobelanguageprogramsbenchmark}, and spans three diverse LM program tasks: classification (\banking{}; \citealt{casanueva2020efficient}), multi-hop claim verification (\hover{}; \citealt{jiang-etal-2020-hover}), and privacy-conscious delegation (\papillon{}; \citealt{siyan2024papillon}).
Each involves different reasoning styles and control flow structures.
Experiments are run using two open-source LMs, \llamathreepointoneshort{}~\citep{grattafiori2024llama} and \qwenthreeshort{}~\citep{yang2025qwen3technicalreport}.

Across these settings, \mmgrpo{} improves performance by \gainsmmgrpovscot{} on average against the model's unadapted reasoning performance.
While \mmgrpo{} does not always surpass the prompt optimized programs via MIPROv2~\citep{opsahl-ong-etal-2024-optimizing}, it complements them effectively: staging MIPROv2 and \mmgrpo{}---\`{a} la BetterTogether---consistently yields higher performance than either method alone, improving by \gainsbettertogethervspo{} and \gainsbettertogethervsmmgrpo{} compared to MIPROv2 and \mmgrpo{}, respectively; and by \gainsbettertogethervscot{} compared to the model's unadapted reasoning performance.
These findings suggest that policy gradient RL and PO offer complementary benefits for LM program training, and we advocate for future work exploring their integration in offline and online settings.

\section{Preliminaries}
\label{sec:preliminaries}

GRPO is an online policy gradient method for LM fine-tuning that operates over \emph{groups} of trajectories, conventionally sharing the \textit{same input prompt}.
The GRPO objective encourages the current policy \( p_{\theta_{\text{old}}} \), parameterized by LM weights \( \theta_{\text{old}} \), to upweight relatively high-reward completions within a group, while applying PPO-style clipping and KL divergence regularization to ensure stable updates.
This results in an updated policy \( p_{\theta} \).

GRPO also makes use of a reference policy \( p_{\theta_{\text{ref}}} \) in the KL divergence penalty, seeking to prevent the updated policy from drifting too far from its original distribution.
Here, we express the original GRPO objective in \autoref{eq:grpo_loss} in terms of prompt--output--reward triples \( (q, o_i, r_i) \), which will let us describe how LM-program traces can be converted into multi-module GRPO training data.

\begin{align}
    \mathcal{J}_{\text{GRPO}}\!
    & \left(\theta\right)
    = \mathbb{E}_{
        \{(q,\, o_i,\, r_i)\}_{i=1}^G
    }
    \notag \\
    &\;\frac{1}{G}\sum_{i=1}^{G}\frac{1}{|o_i|}
    \sum_{t=1}^{|o_i|}
    \Bigg\{
    \min\!\Big(
        \omega_{i,t}\hat{A}_i,\;
        \operatorname{clip}\!\left(
            \omega_{i,t},\,
            1{-}\epsilon,\,
            1{+}\epsilon
        \right)\hat{A}_i
    \Big)
    \notag \\[0.3em]
    &\phantom{\;\frac{1}{G}\sum_{i=1}^{G}\frac{1}{|o_i|}\sum_{t=1}^{|o_i|}\Bigg\{}
    - \beta\,\mathbb{D}_{\mathrm{KL}}\!\Big[
        p_\theta
        \;\|\;
        p_{\theta_\text{ref}}
    \Big]
    \Bigg\},
    \label{eq:grpo_loss}
\end{align}
where $\theta$ indicates the parameters for an LM shared by all groups,
$\hat{A}_i$ is derived from the observed reward $r_i$ (below), and
\[
\omega_{i,t} = \frac{
    p_\theta(o_{i,t} \mid q,\, o_{i,<t})
}{
    p_{\theta_{\text{old}}}(o_{i,t} \mid q,\, o_{i,<t})
}
\]

Each GRPO group is defined as a set of triples \( \mathcal{G} = \{ (q, o_i, r_i) \}_{i=1}^{G} \), constructed by first sampling a fixed prompt from a distribution of questions \( q \sim P(Q) \), and then generating a batch of \( G \) completions \( \{ o_i \}_{i=1}^{G} \sim p_{\theta_{\text{old}}}(O \mid q) \) from the current policy.
Finally, a scalar reward $r_i$ for each $o_i$ is computed with a reward function.
The term \( \omega_{i,t} \) denotes the importance sampling ratio between the new and old policies for the \( t \)th token in a given output.
The scalar reward $r_i$ is then normalized within the group to compute an advantage \( \hat{A}_i \) in the \textit{outcome supervision} formulation of GRPO,
\begin{align}
    \hat{A}_i &= \frac{r_i - \text{mean}(\mathcal{R})}{\text{std}(\mathcal{R})}, \quad
  \mathcal{R} = \{ r_i \}_{i=1}^G,
    \label{eq:grpo_outcome_supervision}
\end{align}
which is applied uniformly across all tokens \( t \) in the corresponding completion \( o_i \), as shown in \autoref{eq:grpo_outcome_supervision}.

\paragraph{LM program formulation}
An LM program \( \Phi \) is composed of LM modules and other tools orchestrated by its control flow.
Let \( \mathcal{M} = \{ M_1, \ldots, M_{|\mathcal{M}|} \} \) denote the set of LM modules used therein, each of which communicates via natural language.

Given a structured input \( x \) (for example, a record with fields such as \texttt{question} or \texttt{document\_titles}), executing \( \Phi(x) \) orchestrates module invocations, transforming inputs and routing outputs between modules.
In other words, \( \Phi(x) \) defines a distribution from which we can sample \( y, \rho \) pairs, where \( y \) is the final output and \( \rho \) is the trajectory of module calls:
\begin{align}
    (y, \rho) \sim \Phi(x), \quad \rho = [\zeta_1, \zeta_2, \dots, \zeta_{|\rho|}]
    \label{eq:sampling_from_lm_program}
\end{align}
Here, the trajectory \( \rho \) records the sequence of module calls, and each trace \( \zeta_t = \langle M_t, q_t, o_t \rangle \) captures the module identity as well as the module-level inputs and outputs at module invocation \( t \) within the program trajectory.
The trajectory \( \rho \) logs only the LM-level calls in their execution order and omits any other control logic. This trace is the same execution record exposed by DSPy programs for prompt optimization, and we reuse it in this work as the basic data structure to conduct online RL weight updates over unchanged LM programs.

Each module \( M \in \mathcal{M} \) is parameterized by a prompt template \( \pi_{M} \) and LM weights \( \theta_{M} \), and may be invoked multiple times.
For example, in a multi-hop setting, a module used to generate new queries could be invoked once for each hop, leading to multiple invocations of that module within a single trajectory $\rho$.
During execution at module invocation \( t \), the prompt template \( \pi_{M_t} \) transforms the input \( q_t \) into a materialized prompt, where we use the same symbols for the structured input \( q_t \) and its materialized prompt to simplify the notation:
\begin{align}
    q_t \gets \pi_{M_t}(q_t)
\end{align}

This prompt is then passed to an LM parameterized by \( \theta_{M_t} \), which samples an output \( o_t \), returned to the control flow of $\Phi$ for subsequent steps:
\begin{align}
    o_t \sim p_{\theta_{M_t}}(\cdot \mid q_t)
\end{align}

This modularity offers several benefits over ReAct-style agents, where every step is appended to a single growing prompt.
Such approaches often inflate contexts and leak unnecessary information across steps.
Further, they are more difficult to test as the smaller subtasks the agent must execute, such as query generation, cannot be easily evaluated independent of the other subtasks.
In contrast, control flow within LM programs specifies exactly which inputs each module sees.
This allows repeated module invocations without carrying along unnecessary context, and further allows for each module to be tested in isolation.
This explicit modularity makes execution more interpretable and is a core reason why recent and concurrent \textit{multi-step} GRPO implementations, which assume a single expanding prompt, are not directly applicable to multi-module LM programs~\citep{jin2025searchr1trainingllmsreason, zeng2025reinforcingmultiturnreasoningllm, wang2025ragenunderstandingselfevolutionllm}.

\paragraph{LM program optimization}

Let $\mathcal{D} = \{(x, m)\}$ be a dataset of inputs $x$ and optional metadata $m$ (e.g., final answer, documents to retrieve, or PII to redact). The goal is to learn the parameters of the given LM program \( \Phi \), namely, the prompt templates \( \pi_M \) and LM weights \( \theta_M \) for each module \( M \in \mathcal{M} \), such that we maximize the expected reward:
\begin{align}
    \mathbb{E}_{
        (x, m) \sim \mathcal{D};\ (y,\rho)\sim\Phi_{\Pi,\Theta}(x)
    }
    \left[
        \mu\left(y, \rho, x, m \right)
    \right]
    \label{eq:lm_program_objective}
\end{align}

Here, the reward function \( \mu(y, \rho, x, m) \) scores the execution, typically based on the final output $y$'s correctness.
Any metadata \( m \) (e.g., gold answers) is not visible to the program during execution but may be used by \( \mu \) for evaluation.
Prompt optimization methods operate on this objective by updating the prompt templates \(\Pi=\{\pi_M\}\), while \mmgrpo{} updates the LM weights \(\Theta=\{\theta_M\}\) using rewards from the same program executions. The BetterTogether framework composes these two choices by first optimizing prompts and then optimizing weights, or alternating both.

\section{Applying GRPO to multi-module LM programs}
\label{sec:method}

\begin{figure*}[tp]
    \centering
    \includegraphics[width=0.84\textwidth]{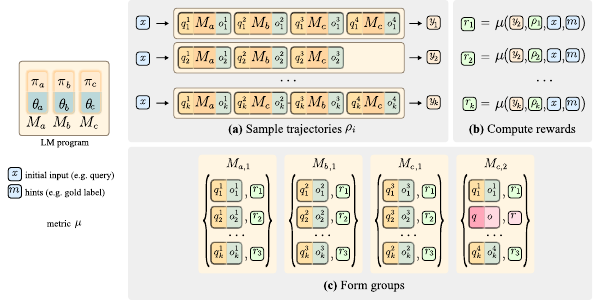}
    \caption{Illustration of \mmgrpo{}'s group building strategy for the module-level variant. We assume an LM program with modules $M_a$, $M_b$, and $M_c$, an input $x$ (with optional hints $m$), and a final reward metric $\mu$. Module-level \mmgrpo{} constructs GRPO groups in three steps: \textbf{(a)} Sample $k$ trajectories by running the LM program. For each input, the program logic selects the next module to call and prepares its input $q_1$. After the module returns an output $o_1$, the program logic prepares the next module input, and so on. Because control flow is governed by the program logic, the sequence and number of module calls can differ across trajectories, even for the same $x$. Each run ultimately produces a final output $y$. \textbf{(b)} A reward function scores the final output (and optionally the trajectory), producing a scalar reward $r$. \textbf{(c)} Module-level \mmgrpo{} then forms GRPO groups by collecting all call data for a given module and relative invocation index (e.g., ``second call to $M_c$'') into a separate group. When a trajectory does not contain a particular module invocation, a padding strategy (e.g., repeating one of the other datapoints in the group with its associated rewards) is applied so that group sizes remain consistent across trajectories.}
    \Description{A three-panel diagram illustrating module-level mmGRPO's group-building strategy for a program with modules M_a, M_b, and M_c. Panel (a) shows k sampled trajectories of program execution; each trajectory is a sequence of module calls that may differ in length and module ordering. Panel (b) shows a reward function evaluating each trajectory's final output and assigning a scalar reward. Panel (c) shows module-level GRPO groups formed by aligning calls across trajectories by (module, invocation index); missing positions are filled by a padding strategy so each group has the same size.}
    \label{fig:mmgrpo_groups}
\end{figure*}

We design our multi-module GRPO framework \mmgrpo{} to be interchangeable with prompt optimizers operating over the same LM program and to rely on the same structure of execution traces that LM programming frameworks like DSPy~\citep{khattab2024dspy} already collect.
As a result, learning algorithms can be swapped without modifying the program itself or entangling the program design process with knowledge of the details of the learning algorithm.

Given a dataset \( \mathcal{D} \) and a reward function \( \mu \), our goal is to optimize an LM program \( \Phi \) consisting of modules \( \mathcal{M} \) by updating the weights \( \theta_{M_i} \) of each module.
In standard GRPO, each group contains completions from a single auto-regressive LM call---i.e., one prompt and its full output.
However, LM programs typically comprise multiple modules with each one invoking its own LM using a custom prompt.
This raises the practical question of how to convert LM-program rollouts into GRPO training groups.
We explore two variants of \mmgrpo{}, with choices that allow our implementation to remain largely modular with respect to existing GRPO implementations.
The main variant we test is module-level \mmgrpo{}, where groups are formed at the module invocation level, but we also test a trajectory-level variant.

\begin{algorithm}[t]
    \caption{\mmgrpo{}: GRPO for multi-module LM programs}
    \label{alg:mmgrpo_main}
    \small

    \begin{algorithmic}[1]
        \Require Student program $\Phi$ with module set $\mathcal{M}$;
        training set $\mathcal{D}$; metric $\mu$; steps $N_{\text{steps}}$;
        group size $G$; batch size $B$; number of example rollouts $K$;
        training hyper-parameters $\Psi_{\text{train}}$

        \Function{mmGRPO}{$\Phi, \mathcal{D}, \mu, N_{\text{steps}}, G, B, K, \Psi_{\text{train}}$}
            \For{$\text{step} = 1$ to $N_{\text{steps}}$}  \label{line:mmgrpo_training_loop}
                \State $\mathcal{B} \gets \textsc{SampleBatch}(\mathcal{D}, B)$  \label{line:mmgrpo_sample_batch}
                \For{each $(x, m) \in \mathcal{B}$}
                    \State $\mathcal{R} \gets \textsc{SampleRollouts}(\Phi, K, x)$  \label{line:mmgrpo_sample_rollouts}
                    \State $\texttt{groups}, \Theta \gets \textsc{FormGroups}(\Phi, \mathcal{R}, G, \mu, x, m)$  \label{line:mmgrpo_form_groups}
                    \For{each $(\mathcal{G}, \theta_M) \in \textsc{Zip}(\texttt{groups}, \Theta)$}
                       \State Update $\theta_M$ via $\mathcal{J}(\mathcal{G})$ with hyper-parameters $\Psi_{\text{train}}$  \label{line:mmgrpo_update}
                    \EndFor
                \EndFor
            \EndFor
            \State \Return $\Phi$ with unchanged prompt templates $\{\pi_{M_i}\}_{i=1}^{|\mathcal{M}|}$
            \Statex \hspace{4.5em}and updated LM weights $\{\theta^*_{M_i}\}_{i=1}^{|\mathcal{M}|}$
        \EndFunction

        \Function{SampleRollouts}{$\Phi, K, x$}
            \State $\mathcal{R} \gets \emptyset$
            \For{$k = 1$ to $K$}
                \State $(y, \rho) \sim \Phi(x)$
                \State $\mathcal{R} \gets \mathcal{R} \cup \{(y, \rho)\}$
            \EndFor
            \State \Return $\mathcal{R}$
        \EndFunction

        \addvspace{0.5em}
        \Statex \textbf{Module-level variant:} $\mathcal{J} = \mathcal{J}^{\text{module}}_{\text{mmGRPO}}$ (\autoref{eq:mmgrpo_loss});\, \textsc{FormGroups} is set to \textsc{FormModuleLevelGroups} (\autoref{alg:mmgrpo_form_module_level_groups}).
        \Statex \textbf{Trajectory-level variant:} $\mathcal{J} = \mathcal{J}^{\text{trajectory}}_{\text{mmGRPO}}$ (\autoref{eq:mmgrpo_loss_trajectory});\, \textsc{FormGroups} groups full trajectories.
    \end{algorithmic}
\end{algorithm}

The main algorithm for \mmgrpo{} is shown in \autoref{alg:mmgrpo_main}, covering both variants we test.
Similar to GRPO, \mmgrpo{} iterates through $N_{\text{steps}}$ training steps (line~\ref{line:mmgrpo_training_loop}).
At each step it samples a batch of training examples (line~\ref{line:mmgrpo_sample_batch}).
Then, for each example, it samples rollouts via \textsc{SampleRollouts} (line~\ref{line:mmgrpo_sample_rollouts}).
It then calls \textsc{FormGroups} to convert these rollouts into a list of GRPO groups paired with the LM weights they update (line~\ref{line:mmgrpo_form_groups}).
Finally, it applies a GRPO-style update to each group (line~\ref{line:mmgrpo_update}) before moving on to the next example.
The two variants of \mmgrpo{} differ only in their choice of \textsc{FormGroups} and the objective $\mathcal{J}$ as specified at the bottom of the algorithm, but the surrounding control flow is shared otherwise.

We now explain how the module-level and trajectory-level variants each instantiate \textsc{FormGroups} and the objective $\mathcal{J}$.

\paragraph{Forming module-level groups}
In the module-level variant, \mmgrpo{} forms groups using the \textsc{FormModuleLevelGroups} algorithm shown in \autoref{alg:mmgrpo_form_module_level_groups}.
We describe it at a high level in this section, with a more detailed treatment saved for \autoref{app:extended_method_module_level_groups}.

\textsc{FormModuleLevelGroups} aligns module calls across trajectories based on both the module identifier and the relative order in which it appears in the trajectory.
This alignment yields module-level GRPO groups, each of the form \( \{ (q_i, o_i, r_i) \}^{G}_{i=1} \), where \( q_i \) and \( o_i \) are extracted from a group of aligned traces all generated by a specific module \( M \), and \( r_i \) is set to the corresponding program-level output reward for the trajectory that generated each trace.

In practice, not all trajectories generated by \( \Phi \) given the same program-level input \( x \) follow the same structure; the program logic may diverge (e.g., by invoking different modules or terminating early), or errors such as module-level parsing failures may halt execution.
To create groups from trajectories with different structures, \textsc{FormModuleLevelGroups} optionally pads smaller groups to a fixed size before applying the loss.
This is exemplified in \autoref{fig:mmgrpo_groups}, where a module-level group with fewer entries than the target size is filled in via the padding strategy.
Refer to \autoref{app:extended_method_module_level_groups} for more details on \textsc{FormModuleLevelGroups} and the padding modes.

Once the groups are formed, the module-level \mmgrpo{} loss in~\autoref{eq:mmgrpo_loss} is applied independently to each module-level group, with two key differences from the original GRPO objective (\autoref{eq:grpo_loss}).
\begin{align}
    \mathcal{J}^{\text{module}}_{\text{mmGRPO}} \!
    & \left(\colorbox{HighlightColorTheta}{$\theta_M$}\right)
    = \mathbb{E}_{
        \{(\colorbox{HighlightColorQ}{$\scriptstyle q_i$},\, o_i,\, r_i)\}_{i=1}^G
    }
    \notag \\[0.3em]
    &\;\frac{1}{G}\sum_{i=1}^{G}\frac{1}{|o_i|}
    \sum_{t=1}^{|o_i|}
    \Bigg\{
    \min\!\Big(
        \omega_{i,t}\hat{A}_i,\;
        \operatorname{clip}\!\left(
            \omega_{i,t},\,
            1{-}\epsilon,\,
            1{+}\epsilon
        \right)\hat{A}_i
    \Big)
    \notag \\[0.3em]
    & \phantom{\;\frac{1}{G}\sum_{i=1}^{G}\frac{1}{|o_i|}\sum_{t=1}^{|o_i|}\Bigg\{}
    - \beta\,\mathbb{D}_{\mathrm{KL}}\!\Big[
        \colorbox{HighlightColorTheta}{$\scriptstyle p_{\theta_M}$}
        \;\|\;
        \colorbox{HighlightColorTheta}{$\scriptstyle p_{\theta_{M_\text{ref}}}$}
    \Big]
    \Bigg\},
    \label{eq:mmgrpo_loss}
\end{align}
where $\colorbox{HighlightColorTheta}{$\scriptstyle\theta_M$}$ are the LM weights for module $M$,
$\hat{A}_i$ is computed from $r_i$ via \autoref{eq:grpo_outcome_supervision}, and
\begin{equation}
    \omega_{i,t} = \frac{
        \colorbox{HighlightColorTheta}{$p_{\theta_M}$}
        (o_{i,t} \mid \colorbox{HighlightColorQ}{$q_i$},\, o_{i,<t})
    }{
        \colorbox{HighlightColorTheta}{$p_{\theta_{M_\text{old}}}$}
        (o_{i,t} \mid \colorbox{HighlightColorQ}{$q_i$},\, o_{i,<t})
    }
    \label{eq:importance_weight}
\end{equation}
First, each group contributes gradients through the LM weights used by the corresponding module: in the general case, modules may have separate weights or adapters.
In our experiments, all trainable modules share one LM adapter, so all module-level groups update the same underlying weights.
Second, unlike GRPO where completions share a single prompt, datapoints in a module-level group may have different prompts \( q_i \), reflecting variation in upstream context.
We use the module-level \mmgrpo{} variant presented here for our main experiments in \autoref{sec:experiments}.

\paragraph{Forming trajectory-level groups}
A natural alternative to module-level grouping is to keep each full program rollout as the unit of GRPO training.
For a trajectory \(\rho_i\), let \(o_{i,j}\) denote the output of the \(j\)-th LM invocation in the trajectory, \(q_{i,j}\) its local prompt, and \(o_{i,j,t}\) the \(t\)-th generated token of that module output.
Let $L_i = \sum_{j=1}^{|\rho_i|} |o_{i,j}|$
denote the total number of generated LM tokens in trajectory \(\rho_i\).
For the trajectory-level variant we evaluate, all trainable module calls share one LM adapter, so we write a single set of weights \(\theta\).
With separate module weights, the corresponding module parameters would be used for the tokens produced by each module.

Omitting clipping and KL terms for presentation, the trajectory-level objective can be written as
\begin{align}
    \mathcal{J}^{\text{trajectory}}_{\text{mmGRPO}}(\theta)
    =
    \mathbb{E}_{\{\rho_i\}_{i=1}^G}
    \frac{1}{G}\sum_{i=1}^{G}
    \frac{1}{L_i}
    \sum_{j=1}^{|\rho_i|}
    \sum_{t=1}^{|o_{i,j}|}
    \left\{
        \omega_{i,j,t}\hat{A}_i
    \right\},
    \label{eq:mmgrpo_loss_trajectory}
\end{align}
where
\begin{align}
    \omega_{i,j,t}
    =
    \frac{
        p_{\theta}(o_{i,j,t} \mid q_{i,j},\, o_{i,j,<t})
    }{
        p_{\theta_{\mathrm{old}}}(o_{i,j,t} \mid q_{i,j},\, o_{i,j,<t})
    }.
\end{align}

\paragraph{Module-level vs. trajectory-level grouping}
The two variants differ in how the same rollout data is grouped and normalized.
Module-level \mmgrpo{} splits trajectories into local prompt--output--reward triples and forms groups from aligned module invocations.
In contrast, trajectory-level \mmgrpo{} keeps each full execution as the grouped unit, before applying the program-level reward to all generated LM tokens in that execution.
Thus, module-level grouping normalizes each module invocation by its own output length, while trajectory-level grouping normalizes by the total generated length of the full rollout. When trajectories differ in structure, module-level grouping also requires alignment and padding choices, whereas trajectory-level grouping operates directly over the full rollouts.

We use the module-level formulation in our main experiments for two practical reasons. First, it maps relatively easily onto existing prompt--completion GRPO trainers: each training example remains a local module prompt, module output, and program-level reward.
Second, because module-level normalization scales each invocation by its own output length, we might expect it to be more suitable when different parts of a program generate outputs of very different sizes.
Trajectory-level grouping is also natural and bakes in fewer assumptions specific to LM programs. We test it as another instantiation in \autoref{sec:experiments}.

\paragraph{Sampling with programs other than the student program}
All of our experiments use an on-policy setting, where we use the student program $\Phi$ to sample rollouts.
One modification that would be simple to implement in the \mmgrpo{} framework is sampling from \textit{teacher programs} that share the same structural interface as the student program (i.e., operating over the same module-level input and output fields), but may differ in their prompt templates (e.g., alternative instructions or few-shot examples) or LM weights (e.g., larger models).
In particular, \textsc{SampleRollouts} can be modified to accept these teacher programs; and then, rather than sampling all rollouts from the student alone, \mmgrpo{} could sample from a configurable mixture of teacher programs.
These variations could enable partially off-policy training, providing greater flexibility in guiding learning with higher-performing policies.
We leave this exploration to future work.

\section{Composing online RL with prompt optimization via BetterTogether}
\label{sec:method_bettertogether}

BetterTogether~\citep{soylu-etal-2024-fine} demonstrates that combining prompt optimization (PO) with weight optimization yields stronger results than using either technique alone, specifically in the context of offline RL via rejection fine-tuning on outcome-filtered trajectories.
Rather than applying weight optimization directly to an unmodified program, the authors first optimize the program's prompt templates and then apply weight optimization on the resulting prompt-optimized program.

This ordering is natural for LM programs because prompt optimization and weight optimization act on different elements of the same program. Prompt optimization changes the instructions, demonstrations, and module interfaces that shape the program's rollouts at a high level of abstraction. \mmgrpo{} then updates the LM weights under the rollout distribution induced by those prompts, using the same program-level metric.

We extend the BetterTogether approach to the online RL setting for the first time using \mmgrpo{}.
In particular, we experiment with combining the module-level variant of \mmgrpo{} with a strong prompt optimizer, MIPROv2 \citep{opsahl-ong-etal-2024-optimizing}.
While \citet{soylu-etal-2024-fine} also experiment with alternative compositions, such as running prompt optimization after weight tuning, in this work we focus on the former: applying an \mmgrpo{} variant to a program on top of automatic prompt optimization.

\section{Experiments}
\label{sec:experiments}

\subsection{LMs and datasets}
\label{sec:experiments_lms_and_datasets}

For both LM programs and datasets, we start with implementations from the Language Programs Benchmark (LangProBe), which was created to allow easy evaluation of optimization approaches for multi-module LM programs \citep{tan2025langprobelanguageprogramsbenchmark}.
The DSPy implementation of these LM programs for each task is shown in \autoref{app:task_details}.

\textbf{LMs.} We run our experiments on \llamathreepointoneshort{}~\citep{grattafiori2024llama} and \qwenthreeshort{}~\citep{yang2025qwen3technicalreport}.
Although \mmgrpo{} allows for different LM copies to learn separate weight updates for the different modules of a program, we focus on the multi-task setting where a single set of weights is shared for all modules in an LM program.

\textbf{Classification.}
\banking{} is an intent classification benchmark involving $13{,}083$ labeled customer service queries from the banking domain \citep{casanueva2020efficient}.
The task is to assign each user query to one of 77 intent classes, an example of which is shared below.

\begin{datapoint}{Example datapoint for \banking{}}
    \label{datapoint:banking}
    \textbf{text:}
    I have been waiting over a week. Is the card still coming?

    \addvspace{\smallskipamount}
    \textbf{label:}
    card\_arrival
\end{datapoint}

We implement a simple program for this task using a single Chain-of-Thought (CoT) module \citep{wei2023chainofthoughtpromptingelicitsreasoning}, which first produces a reasoning trace before predicting the intent label.
For evaluation we compute the exact match between the ground-truth label and the generated label.
We intentionally include this single-module program as a sanity check.
With only one module, both \mmgrpo{} variants reduce to standard GRPO, allowing us to verify that the algorithm recovers GRPO performance in the simplest case before evaluating it on more complex multi-module systems.

For training and evaluation, we randomly sample $250$ training examples and $500$ for evaluation.

\textbf{Privacy-conscious delegation.}
The Private User Prompt Annotations (PUPA) benchmark by \citet{siyan2024papillon} focuses on privacy-preserving question answering, where the goal is to respond to user queries without exposing private information to external APIs.

\begin{datapoint}{Example datapoint for \papillon{}}
    \label{datapoint:papillon}
    \textbf{user\_query:}
    Send a friendly e-mail opening to Sara telling her thank you for calling me back so quickly, taking the time to discuss with me FRC's IPAEP, and for your immediate willingness to make reasonable accommodations to help break down the barriers caused by my disabilities so that I can have equal access to your program. Your kindness and compassion is appreciated.

    \addvspace{\smallskipamount}
    \textbf{target\_response:}
    Dear Sara, I hope this email finds you well. I just wanted to take a moment to express my sincere gratitude for your prompt response to my call and the time you took to discuss FRC's IPAEP with me. I truly appreciate your immediate willingness to make reasonable accommodations to ensure that I have equal access to your program and to help break down the barriers caused by my disabilities...

    \addvspace{\smallskipamount}
    \textbf{pii\_str:}
    sara\textbar\textbar ipaep\textbar\textbar frc
\end{datapoint}

We use \papillon{}, also from \citet{siyan2024papillon}, a two-module pipeline that generates a redacted version of a private user query, sends the redacted query to an untrusted but more performant external model, then uses the response of that external model to generate the final response.
\papillon{} tests \mmgrpo{} on a multi-module system whose modules play structurally distinct roles, redaction and response synthesis, each invoked once per trajectory.

We use \gptfouronemini{}~\citep{openai_gpt4_1mini} as the external LM.
As described in \citet{siyan2024papillon}, the evaluation metric is a composite score which takes into account the content of the response and the amount of private information that was leaked, both of which are scored by a large LM acting as a judge.
We evaluate this setup using $111$ training examples and $221$ for evaluation.

\textbf{Multi-hop claim verification.} \hover{}~\citep{jiang-etal-2020-hover} is a claim verification benchmark where the task is to extract facts from multiple relevant Wikipedia articles and decide whether a given claim is supported.

\begin{datapoint}{Example datapoint for \hover{}}
    \label{datapoint:hover}
    \textbf{claim:}
    This director is known for his work on Miss Potter. The Academy of Motion Picture Arts and Sciences presents the award in which he was nominated for his work in "Babe".

    \addvspace{\smallskipamount}
    \textbf{titles:}
    ['Miss Potter', 'Chris Noonan', 'Academy Award for Best Director']
\end{datapoint}

The claims in \hover{} are \textit{multi-hop} in that they require multi-hop reasoning by connecting information found in different articles.
The original dataset has $18{,}171$ train and $4{,}000$ development and test examples derived from the examples in the HotPotQA dataset \citep{yang-etal-2018-hotpotqa}.
Our program for \hover{} consists of 2 modules, a query generation module and a fact summarization module, called iteratively over 4 hops, along with a \colberttwo{}~\citep{santhanam2022colbertv2effectiveefficientretrieval} retriever indexed on the short snippets from the Wikipedia (2017) dump provided with the HotPotQA dataset, shared with \hover{}.
This program tests \mmgrpo{} along a different axis than \papillon{} or \banking{}, as multiple modules are invoked repeatedly across many hops within a single trajectory.

We refer to the particular 4-hop variant Hover program we use as \hoverfour{}, in order to differentiate it from the one provided in \citet{tan2025langprobelanguageprogramsbenchmark}.
The program returns up to 100 passages at the end, and the final metric evaluates whether the gold passages are found within the returned passages using Recall@100.
We build our splits from the original train split, randomly sampling $500$ examples each for our train and evaluation splits, while ensuring that we do not sample any two examples derived from the same HotPotQA question.

\subsection{Baseline and method details}
\label{sec:experiments_baseline_and_details}

We evaluate each of our LM and task pairs with vanilla Chain-of-Thought (CoT) and a prompt optimizer, to serve as baselines.
We demonstrate our module-level \mmgrpo{} optimizer in two configurations: \mmgrpo{}, and BetterTogether \mmgrpo{}.
While each method assumes access to the same program-level evaluation metric, none uses intermediate module-level supervision. Instead, training data is generated online by running the program and scoring full program executions with the metric.
We build on the DSPy framework \citep{khattab2024dspy} to run our baseline experiments and develop our new \mmgrpo{} optimizers, and draw inspiration from the Verifiers library \citep{brown2025verifiers}.

\textbf{Inference.} We use the \texttt{vLLM}~\citep{kwon2023efficient} engine for sampling with max context length of $32{,}768$ tokens for inference.
We set max tokens to $1{,}032$ and re-try each query up to $3$ times in case of module parsing errors.
For \qwenthreeshort{}, we use $\mathrm{sampling\_temperature}=0.6$, $\mathrm{top\_p}=0.95$ and $\mathrm{top\_k}=20$ following the parameters used for its instruction training as noted in the official technical report \citet{yang2025qwen3technicalreport}.
For \llamathreepointoneshort{}, we use $\mathrm{sampling\_temperature}=0.6$ and $\mathrm{top\_p}=0.9$ following the official model card's generation configuration in HuggingFace~\citep{meta_llama_3_1_8b_instruct_2024}.

\textbf{Vanilla CoT.} We adopt the Chain-of-Thought (CoT) prompting method introduced by \citet{wei2023chainofthoughtpromptingelicitsreasoning}, where each module's prompt instructs the language model to first generate a \textit{reasoning} field before producing its final output (e.g., answer, search\_query).
This creates a strong prompt baseline to compare our methods against.
Unless stated otherwise, both the prompt-optimization and \mmgrpo{} methods described below begin training from this base CoT prompt.
We refer to this initial prompt configuration as the ``Vanilla CoT'' program.

\textbf{MIPROv2.} We use the state-of-the-art prompt optimizer Multiprompt Instruction PRoposal Optimizer Version 2 (MIPROv2; \citealt{opsahl-ong-etal-2024-optimizing}) as our prompt-optimized baseline.
MIPROv2 jointly optimizes module prompts—free-form instructions and optional few-shot examples—by (i) bootstrapping candidate exemplars, (ii) proposing instructions, and (iii) selecting high-yield instruction–example combinations via Bayesian optimization over program-level metrics.
For our experiments, we use the \texttt{medium} setting, which uses \texttt{12} trials; \texttt{12} few-shot and \texttt{6} instruction candidates, and automatically uses 80\% of the train set for validation.
We refer to the program optimized using MIPROv2 with these settings as the prompt-optimized program and re-use it for the BetterTogether strategy below.

\textbf{mmGRPO.}
Except when specified otherwise, \mmgrpo{} refers to the module-level variant used in our main experiments.
We train our models using the \texttt{GRPOTrainer} in HuggingFace TRL \citep{vonwerra2020trl}, each with a maximum context length of $8{,}192$ tokens.
Training is performed with a temperature of $0.6$, a learning rate of $1 \times 10^{-5}$, gradient accumulation steps of $20$, with per device train batch size of $1$.
We use $\beta = 0.01$ and gradient norm clipping of $0.1$ for \qwenthreeshort{}; and $\beta = 0.04$ and gradient norm clipping of $0.5$ for \llamathreepointoneshort{}.

We run \mmgrpo{} for 750 steps, using 4 training examples per step.
At each step, we randomly draw 4 examples from the training dataset.
For each example, we generate 12 rollouts, which are then grouped into module-level or trajectory-level GRPO groups depending on which \mmgrpo{} variant is used.
We use a train context length of $8{,}192$ tokens, which is used to filter any trajectory with a module-level prompt and completion longer than this.
We apply Low-Rank Adaptation (LoRA, \citealt{hu2021loralowrankadaptationlarge}) with rank $\mathrm{r}=16$, $\mathrm{lora\_alpha}=64$, $\mathrm{lora\_dropout}=0.05$, targeting the projection modules $[\mathrm{q}, \mathrm{k}, \mathrm{v}, \mathrm{o}, \mathrm{up}, \mathrm{down}, \mathrm{gate}]$.
We run all of our \mmgrpo{} experiments below using these same settings.

\textbf{mmGRPO with BetterTogether.}
We further evaluate a setting that combines prompt optimization with the weight optimization of \mmgrpo{} following the BetterTogether framework \citep{soylu-etal-2024-fine}.
For these experiments, we only use the module-level \mmgrpo{} variant.
Specifically, instead of optimizing the LM program's weights directly, we first apply prompt optimization to identify high quality prompts for each module.
The prompts are then kept fixed, and \mmgrpo{} is used to optimize the program's weights.
We refer to this configuration as BetterTogether(PO, \mmgrpo{}), which is the main compositional setting studied in our experiments.

\subsection{Main results}
\label{sec:experiments_main_results}

\begin{table*}[tp]
    \centering
    \setlength{\tabcolsep}{7pt}  %
    \caption{
    Performance of different learning algorithms across three LM programs: a single-stage intent-classification program, \banking{}, two-stage privacy-conscious delegation program, \papillon{}, and a four-stage retrieval based question answering program \hoverfour{}. MIPROv2 represents a prompt optimization baseline where prompts for all the stages in a program are jointly optimized, while Vanilla CoT refers to vanilla chain-of-thought prompting. Both \mmgrpo{} and MIPROv2 improve over the untuned baseline, though neither consistently dominates the other. The best overall performance is achieved by the BetterTogether variant of \mmgrpo{}, which first applies prompt optimization using MIPROv2 and then fine-tunes using \mmgrpo{}. We report evaluation accuracy for each cell, averaged over 3 seeds.
    }

    \resizebox{\textwidth}{!}{
        \begin{tabular}{@{} lccccccccc @{}}
            \toprule
            \multirow{2}{*}{Strategy} & \multicolumn{2}{c}{\banking{}} & \multicolumn{2}{c}{\papillon{}}     & \multicolumn{2}{c}{\hoverfour{}} & \multicolumn{3}{c}{Avg Scores} \\
            \cmidrule(lr){2-3} \cmidrule(lr){4-5} \cmidrule(lr){6-7} \cmidrule(lr){8-10}
            & \llamathreepointoneshorter{} & \qwenthreeshorter{}
            & \llamathreepointoneshorter{} & \qwenthreeshorter{}
            & \llamathreepointoneshorter{} & \qwenthreeshorter{}
            & \llamathreepointoneshorter{} & \qwenthreeshorter{} & All \\
            \midrule
            \textit{Baseline Strategies}: & & & & & & & & & \\
            \addlinespace[0.3em]
            \quad Vanilla CoT
                & $58.4_{\pm 1.4}$ & $64.6_{\pm 1.2}$
                & $76.2_{\pm 1.3}$ & $78.3_{\pm 0.9}$
                & $59.5_{\pm 2.2}$ & $60.6_{\pm 0.6}$
                & 64.7 & 67.8 & 66.3 \\
            \quad MIPROv2 (PO)
                & $59.4_{\pm 0.4}$ & $65.9_{\pm 1.1}$
                & $83.9_{\pm 5.6}$ & $78.1_{\pm 4.3}$
                & $63.4_{\pm 3.0}$ & $69.3_{\pm 0.8}$
                & 68.9 & 71.1 & 70.0 \\
            \midrule
            Module-level \mmgrpo{} \textit{Strategies}: & & & & & & & & & \\
            \addlinespace[0.3em]
            \quad \mmgrpo{}
                & $\mathbf{63.7}_{\pm 3.2}$ & $64.9_{\pm 0.5}$
                & $83.9_{\pm 0.5}$          & $\mathbf{83.3}_{\pm 1.8}$
                & $60.2_{\pm 1.2}$          & $71.0_{\pm 1.0}$
                & 69.3 & 73.1 & 71.2 \\
            \quad BetterTogether(PO, \mmgrpo{})
                & $\mathbf{63.7}_{\pm 2.6}$ & $\mathbf{69.1}_{\pm 1.8}$
                & $\mathbf{86.5}_{\pm 5.0}$ & $81.1_{\pm 4.1}$
                & $\mathbf{68.3}_{\pm 2.7}$ & $\mathbf{71.5}_{\pm 1.5}$
                & \textbf{72.8} & \textbf{73.9} & \textbf{73.4} \\
            \bottomrule
        \end{tabular}
    }
    \label{tab:main}
\end{table*}

The main results, using the module-level \mmgrpo{} variant, are shown in \autoref{tab:main}.
We share an ablation experiment using the trajectory-level variant in \autoref{sec:experiments_ablation_trajectory}.

\textbf{\mmgrpo{} and BetterTogether(PO, \mmgrpo{}) consistently improve over their respective baselines.}
We can see that the \mmgrpo{} row is consistently higher than the ``Vanilla CoT'' row, \gainsmmgrpovscot{} on average.
Similarly, BetterTogether(PO, \mmgrpo{}) shows consistent gains over the ``MIPROv2 (PO)'' row, \gainsbettertogethervspo{} on average.
These show that \mmgrpo{} is effective at finding better policies for the provided program across all LM--task pairs.

We believe these results are consistent with a key observation about online RL methods, such as \mmgrpo{}, where their effectiveness depends heavily on the quality of the initial rollouts.
When the base policy is too weak, exploration bottlenecks can limit \mmgrpo{}'s ability to discover high-reward trajectories, causing poor gains for complex LM programs such as those used for HoVer.
Instead, prompt optimization improves the reliability of early trajectories, providing a more favorable initialization for weight updates and allowing \mmgrpo{} to exploit higher-quality rollouts.
This also can help explain why BetterTogether consistently surpasses either method alone.

\textbf{\texttt{PO} is competitive with lower computational budgets.}
When averaged across all tasks and models, MIPROv2 alone improved upon the Vanilla CoT strategy by \gainspovscot{} compared to \mmgrpo{}'s \gainsmmgrpovscot{} improvement.
However, MIPROv2 achieved these results significantly faster while using fewer GPU-hours.
On average, our vanilla \mmgrpo{} experiments took 18.7 hours using 2 \gpuhonehundred{} GPUs whereas MIPROv2 took only 1.4 hours on average and only required 1 \gpuhonehundred{} GPU.
These results indicate that PO approaches like MIPROv2 are likely more feasible for settings with lower computation budgets.

\textbf{BetterTogether(PO, \mmgrpo{}) performs the best in most task pairs.}
The BetterTogether(PO, \mmgrpo{}) approach improves over the Vanilla CoT by \gainsbettertogethervscot{}, MIPROv2 by \gainsbettertogethervspo{}, and vanilla \mmgrpo{} by \gainsbettertogethervsmmgrpo{}.
This shows the value of high-quality rollouts at the start of \mmgrpo{} training, as performing PO generates stronger rollouts, leading to a more robust training signal early in the training runs.

\subsection{Trajectory-level \mmgrpo{} experiment}
\label{sec:experiments_ablation_trajectory}

Fully exploring the implementation space of \mmgrpo{} is beyond the scope of this study.
However, to verify that GRPO-style optimization of LM programs is not tied to module-level grouping or other incidental choices, we also experiment with a trajectory-level variant that we have developed since running the main experiments of this paper.
We run it on \hover{} using \qwenthreeshort{}.
This variant groups full program executions rather than aligned module invocations.

We find the trajectory-level \mmgrpo{} performance is strong, and with the usage of the CISPO \citep{minimax2025minimaxm1scalingtesttimecompute} loss and additional tuning of group and batch sizes, we are able to achieve a Recall@100 of $75.3$ compared to the module-level result in \autoref{tab:main} of $71.0$.
This indicates a much larger possible design space of more powerful policy-gradient RL algorithms for multi-module systems to be explored in future work.
We find this particularly exciting with the recent introduction of a new generation of more reflective and more powerful prompt optimizers such as GEPA \citep{agrawal2026gepareflectivepromptevolution} and Meta-Harness \citep{lee2026metaharnessendtoendoptimizationmodel} which could further improve the rich composition we initially explore with our BetterTogether experiments.

\subsection{Qualitative case studies}
\label{sec:qualitative_analysis}

To better understand how \mmgrpo{} changes program behavior, we manually inspect \hover{} examples where the module-level \mmgrpo{} variant achieves higher Recall@100 than both Vanilla CoT and MIPROv2.
The examples below are not intended as a complete causal explanation of the gains in \autoref{tab:main}, but they illustrate two recurring behavioral differences we observed across inputs and rollouts.

\textbf{More retrieval-oriented query generation.}
In several improved cases, \mmgrpo{} changes the \textit{query generation} module from producing sentence-like questions to producing compact semantic retrieval queries.
For example, for a claim involving Jessica Stroup, Alex Wise, Kyle, and the Marvel Cinematic Universe, the Vanilla CoT variant generates a full sentence query: \textit{``Is there a four-part web series featuring Jessica Stroup and Alex Wise as Kyle in the Marvel Cinematic Universe?''}
MIPROv2 instead produces the shorter query \textit{``Alex Wise Iron Fist character''}.
Since the program uses a \colberttwo{} retriever~\citep{santhanam2022colbertv2effectiveefficientretrieval}, this shorter keyword-style query is better matched to the retrieval backend.
The \mmgrpo{} variant produces an even more specific query, \textit{``Alex Wise Jessica Stroup Kyle Iron Fist Marvel Cinematic Universe''}, which retrieves all $3$ relevant documents in this example, compared with $2$ for MIPROv2 and $0$ for Vanilla CoT.

A second case shows improvements for a related failure mode.
For the claim \textit{``The host of the show that featured LA Urban Rangers provided the voice of Backson in an animated feature film.''}, Vanilla CoT refuses to generate a new query, stating that no additional search is needed because the claim is already supported by the current key facts.
MIPROv2 generates a partial query, \textit{``Who was the host of the show that featured LA Urban Rangers''}, while \mmgrpo{} produces the more targeted query \textit{``Huell Howser California's Gold LA Urban Rangers Backson Winnie the Pooh 2011''}.
These qualitative examples suggest that \mmgrpo{} can improve the module's ability to produce queries that are both concise and targeted to the remaining missing evidence.

\textbf{Passing verification state between modules.}
We also observe cases where \mmgrpo{} changes how the \textit{append notes} module uses the \texttt{key\_facts} field.
Rather than only restating facts from the claim, the module sometimes records whether those facts have actually been supported by retrieved passages.

For example, consider the claim \textit{``The player involved in the Helmet Catch with David Tyree was the first overall pick in the 2004 NFL Draft and had the Manning Field named after him.''}
The Vanilla CoT and MIPROv2 variants both include the statement \textit{``The Manning Field was named after Eli Manning''} in \texttt{key\_facts}, even though the retrieved passages have not yet established this fact.
By contrast, the \mmgrpo{} variant writes: \textit{``There is no mention of a Manning Field being named after Eli Manning in the provided search results.''}
This gives the downstream query-generation module a more accurate picture of what remains to be fetched and verified.

A similar pattern appears in another claim involving Guangxi University of Technology and Xiachengzi Railway Station, where \mmgrpo{} writes that Muling is a city in China but that its exact location and area are not specified in the retrieved results.
Again, the module is not merely accumulating candidate facts; it is also tracking gaps in the evidence gathered so far.

Overall, these case studies suggest that \mmgrpo{} can change both local module behavior and cross-module information flow.
The \textit{query generation} module becomes more retrieval-oriented, while the \textit{append notes} module sometimes passes forward a lightweight verification state that helps later queries target missing evidence.

\section{Related work}
\label{sec:related_work}

\textbf{Prompt optimization.}
Much recent work adapts prompts to fit data and tasks, i.e., optimizes the prompts used to invoke an LM rather than its weights.
Broadly, approaches include (i) \emph{instruction generation}, in which LMs synthesize candidate instruction templates~\citep{yang2024large, zhou2023large, pryzant-etal-2023-automatic, fernando2024promptbreeder}; (ii) \emph{gradient-based} methods that optimize discrete text via differentiation (e.g., soft/continuous prompts) and then project back to tokens~\citep{shin-etal-2020-autoprompt, wen2023hard}; and (iii) \emph{RL-based} editors that treat prompt edits as actions and optimize them against trajectory-level rewards~\citep{deng-etal-2022-rlprompt, zhang2023tempera, hao2023optimizing}, among many others.

\textbf{Weight optimization.}
Proximal Policy Optimization (PPO) has been widely used for post-training language models with reinforcement learning, particularly when aligning language models with human preferences or feedback \citep{schulman2017proximal, ouyang2022training}.
In the context of Reinforcement Learning from Human Feedback (RLHF), PPO uses an actor-critic approach where a reward model is trained to approximate human preferences and a policy model is optimized based on predicted rewards from the reward model.
However, PPO is computationally intensive, as it requires training and running both the policy model and the reward model during optimization.

Recently, Direct Preference Optimization (DPO) algorithms have emerged as a simpler alternative that avoids explicit reward modeling and instead learns from contrastive preference pairs \citep{rafailov2023direct}.
Similarly, Group Relative Policy Optimization (GRPO) offers an efficient alternative to PPO by avoiding the need for a value model and instead relying on estimated advantages through relative rewards within a group of rollouts \citep{shao2024deepseekmathpushinglimitsmathematical}.

\textbf{Optimization of LM programs' prompts and weights.} Existing work has explored optimizing LM programs with prompt optimizers, including those that focus primarily on rejection sampling \citep{khattab2024dspy} and others that extend this to use Bayesian optimization for selecting the instruction-demonstration candidates that are most promising \citep{opsahl-ong-etal-2024-optimizing}.
Additional work \citep{soylu-etal-2024-fine} has explored combining weight optimizers with prompt optimizers for additional benefit, but in the context of offline RL.
However, adapting some techniques to LM Programs requires making a number of decisions (\autoref{sec:preliminaries}) and presents substantial implementation challenges.
The present work describes how we generalize online RL via GRPO to LM programs composed of multiple modules.

\section{Conclusion}
\label{sec:conclusion}

We introduce \mmgrpo{}, an implementation of GRPO for multi-module LM programs that propagates rewards across disjoint modules.
Towards maintaining the separation of the programming and learning concerns in LM programs, we design \mmgrpo{} to bolt onto existing programs without modification, making it interchangeable with prompt optimizers.
Our design preserves GRPO's practicality while accommodating heterogeneous prompts and partial trajectories, and our evaluations demonstrate that \mmgrpo{} is highly effective in navigating challenging credit assignment problems without requiring intermediate supervision.
We further show that combining \mmgrpo{} with state-of-the-art prompt optimization methods via BetterTogether yields the strongest overall performance in the majority of settings, revealing a complementary relationship between weight and prompt optimization for online RL methods.

While our experiments demonstrate the promise of multi-module RL formulations, this work has limitations.
We use 8-billion parameter language models, which may not reflect how \mmgrpo{} performs with larger models, and we also rely on LoRA for fine-tuning; while efficient, this may limit training performance compared to full-parameter updates.
Further, we evaluate only two \mmgrpo{} implementations despite many possible alternative formulations.
Finally, while \banking{} is a well-understood classification task, we study it in a limited-feedback setting where models only receive rewards derived from bootstrapped rollouts, not supervised intent labels.
While supervised training enables encoder models to perform well on this task, we focus on investigating whether current methods can achieve strong performance from reward signals on the rollouts alone.
Our results suggest that this is not yet the case.

We acknowledge the use of light assistance from AI tools in the preparation of portions of this paper. All outputs were reviewed, verified, and edited by the authors.

\ifx\papervisibility\visibilityrevealed
    \begin{acks}
    D.S. was partially supported by the Laude Institute, by IBM as a founding member of the Stanford Institute for Human-Centered Artificial Intelligence (HAI), by PwC as a member of HAI, and by the HAI Hoffman–Yee Grant ``Dendritic Computation for Knowledge Systems''.
    N.Z. was partially supported by NSF IIS-2142827, IIS-2234058, and by the Laude Institute.
    L.A. was partially supported by the Laude Institute and the Amazon AI PhD Fellowship.
    D.S. thanks the members of the StanfordNLP group for their support and feedback.
\end{acks}
\fi

\bibliographystyle{ACM-Reference-Format}
\bibliography{references}

\appendix

\onecolumn
\renewcommand{\thetable}{\Alph{section}.\arabic{table}}
\numberwithin{table}{section}

\section*{Appendix}
\section{Forming module-level groups}
\label{app:extended_method_module_level_groups}

We now describe how \mmgrpo{} constructs GRPO-style groups at the module level for LM programs.
Once the rollouts are sampled, \mmgrpo{} constructs \textit{module-level} GRPO groups via the \textsc{FormModuleLevelGroups} function described in \autoref{alg:mmgrpo_form_module_level_groups}.
Each GRPO group is defined as a list of \( G \leq R \) triples \( \{ (q_i, o_i, r_i) \}_{i=1}^G \), where each element consists of a module-level input prompt \( q \), the corresponding output \( o \), and the final trajectory-level reward \( r \).
In practice, one can use \( G < R \), the number of rollouts, to leave room for post-hoc adjustments to group size (discussed later in this section).

\begin{algorithm}
    \caption{\textsc{FormModuleLevelGroups}: Create module-level GRPO groups for \mmgrpo{}}
    \label{alg:mmgrpo_form_module_level_groups}
    \small

    \begin{algorithmic}[1]
        \Require Student program $\Phi$ with module set $\mathcal{M}$; rollouts $\mathcal{R} = \{(y_j, \rho_j)\}_{j=1}^{R}$; group size $G$; metric $\mu$; input $x$; input metadata $m$

        \Function{FormModuleLevelGroups}{$\Phi, \mathcal{R}, G, \mu, x, m$}  \label{line:fmlg_declaration}
            \State $\mathtt{groups\_dict} \gets \textsc{DefaultDict}(\text{list})$ \label{line:fmlg_initialize_groups}
            \For{each $(y, \rho) \in \mathcal{R}$} \label{line:fmlg_rollouts_loop}
                \State $r \gets \mu(y, \rho, x, m)$  \label{line:fmlg_compute_reward}
                \State $\mathtt{counter} \gets \textsc{DefaultDict}(\text{int})$
                \For{each trace $\zeta = (M, q, o)\in \rho$} \label{line:fmlg_trajectory_loop}
                    \State Append $(q, o, r)$ to $\mathtt{groups\_dict}[(M, \mathtt{counter}[M])]$    \label{line:fmlg_append_grpo_group_item}

                    \State $\mathtt{counter}[M] \gets \mathtt{counter}[M] + 1$  \label{line:fmlg_increment_order}  %
                \EndFor
            \EndFor

            \State $\mathtt{groups\_dict} \gets \textsc{PadGroups}(\mathtt{groups\_dict})$ \label{line:fmlg_pad}
            \State $\mathtt{groups} \gets [\textsc{SelectKDiverseElements}(\mathcal{G}, G) \mid \mathcal{G} \in \textsc{Values}(\mathtt{groups\_dict})]$ \label{line:fmlg_diverse_elements}
            \State $\Theta \gets [\theta_M \mid (M, \_) \in \textsc{Keys}(\mathtt{groups\_dict})]$ \label{line:fmlg_extract_theta}
            \State \Return $\mathtt{groups}, \Theta$ \label{line:fmlg_return}
        \EndFunction

        \Statex Assume \textproc{DefaultDict}, \textproc{Keys}, and \textproc{Values} are provided
        \Statex Refer to the prose in \autoref{app:extended_method_module_level_groups} for descriptions of \textproc{PadGroups} and \textproc{SelectKDiverseElements}
        \end{algorithmic}
\end{algorithm}

Given the program \( \Phi \), the list of output--trajectory tuples \( \mathcal{R} \), the desired GRPO group size \( G \), the metric \( \mu \), the input \( x \), and the input metadata \( m \), \textsc{FormModuleLevelGroups} iterates over each output--trajectory pair in \( \mathcal{R} \) (Line~\ref{line:fmlg_rollouts_loop}), computing a corresponding score \( r = \mu(y, \rho, x, m) \) (Line~\ref{line:fmlg_compute_reward}).
If the corresponding trajectory is incomplete, a fallback reward is assigned (e.g., a formatting error penalty).
Following this, it iterates over the traces in each trajectory (Line~\ref{line:fmlg_trajectory_loop}).
Each trace contributes a triple \( (q, o, r) \) consisting of the module-level input, output, and final trajectory reward.
This triple is added to the group corresponding to \( (M, k) \), where \( k \) is the relative invocation index of \( M \) in the trajectory (tracked by $\mathtt{counter}[M]$ in \autoref{alg:mmgrpo_form_module_level_groups}, Line~\ref{line:fmlg_append_grpo_group_item}); $\mathtt{counter}[M]$ is incremented after each occurrence (Line~\ref{line:fmlg_increment_order}).
To ensure uniform group sizes despite variability in module invocation counts across trajectories, Lines~\ref{line:fmlg_pad} and~\ref{line:fmlg_diverse_elements} apply post-processing steps that adjust each group to have exactly \( G \) elements, as detailed later in this section.
Finally, Line~\ref{line:fmlg_extract_theta} constructs a list of LM weight references, one corresponding to each group, and both this list and the final GRPO groups are returned (Line~\ref{line:fmlg_return}).

As a result, \textsc{FormModuleLevelGroups} creates GRPO groups defined by both the module identity and the relative position within the trajectory with respect to the other calls to the same module.
Let \( K_{M_i, \rho_j} \) denote the number of times module \( M_i \) is invoked in trajectory \( \rho_j \) for \( (y_j, \rho_j) \in \mathcal{R} \); then the total number of GRPO groups formed across all trajectories is \( \sum_i \max_j K_{M_i, \rho_j} \), where \( M_i \in \mathcal{M} \) for the given \( \mathcal{R} \).
Each resulting group is a list of module-level \( (q, o, r) \) triples, corresponding to structurally aligned invocations of a given module at a specific position in the trajectory.
In contrast to standard GRPO, which produces a single group per set of rollouts in single-stage settings, \mmgrpo{} yields a list of groups, one for each module and relative invocation position.
To ensure uniform group sizes and handle variation across trajectories, \mmgrpo{} applies two \textit{post-processing} steps: \textsc{PadGroups} and \textsc{SelectKDiverseElements}, described next.

\paragraph{Handling variably invoked trajectories with \textsc{PadGroups}}
If every module \( M_i \) in the student program is invoked the same number of times \( K_{M_i, *} \) across all trajectories \( \rho_j \) where \( (y_j, \rho_j) \in \mathcal{R} \), then each constructed GRPO group will contain exactly \( R \) triples prior to the call to Line~\ref{line:fmlg_pad} in \autoref{alg:mmgrpo_form_module_level_groups}.
For example, suppose the LM program consists of two modules, \( M_1 \) and \( M_2 \), and \( R = 3 \) trajectories are sampled.
If, in every trajectory, the program calls \( M_1 \) exactly twice and \( M_2 \) exactly once, then \mmgrpo{} will form three GRPO groups: two for \( M_1 \) (corresponding to its first and second calls) and one for \( M_2 \).
Each of these groups will contain exactly three triples, one from each trajectory, without requiring any padding or truncation.
This scenario arises when all executions yield structurally identical trajectories and none encounter parsing or runtime errors.

However, in practice, these conditions may not hold: some modules may be invoked fewer times due to variation in control flow, while others may terminate early due to parsing failures or other runtime errors.
In such cases, certain (module, invocation-index) GRPO groups may contain fewer than \( G \) elements.
To address this, \mmgrpo{} applies post-processing strategies to ensure that each group has a uniform size, with a call to the \textsc{PadGroups} function, described here.

The behavior of \textsc{PadGroups} is controlled by a \texttt{padding\_mode} hyper-parameter (not explicitly noted in the function call to it in~\autoref{alg:mmgrpo_main}), which supports two values: \texttt{truncate} and \texttt{fill}.
Under the \texttt{truncate} strategy, it discards all GRPO groups for module \( M_i \) whose invocation index exceeds \( \min_j K_{M_i,\rho_j} \), ensuring that only groups with complete representation across all trajectories are retained.
Under the \texttt{fill} strategy, it pads each trajectory's list of \( M_i \) invocations up to length \( \max_j K_{M_i,\rho_j} \) by uniformly sampling (with replacement) from that trajectory's existing \( M_i \) invocations.
We use the \texttt{fill} setting for the experiments reported in this paper.

\paragraph{Ensuring diversity in groups with \textsc{SelectKDiverseElements}} After standardizing group sizes across trajectories, \mmgrpo{} further adjusts each group to ensure it contains exactly \( G \) elements, the target GRPO group size.
Rather than sampling elements uniformly at random, it invokes the \textsc{SelectKDiverseElements} function, which selects (or duplicates) elements to form a group of size \( G \) while maximizing diversity within the group.
This function handles both down-sampling (when the group has more than \( G \) elements) and up-sampling (when it has fewer), favoring selections that increase reward variance in the sampled prompt-output pairs.
Contemporaneously, \citet{xu2026rolloutsusefuldownsamplingrollouts} propose a similar variance-based selection strategy, demonstrating that promoting diversity in GRPO groups improves held-out generalization.

\section{Task details}
\label{app:task_details}

The DSPy implementations for the LM programs for \banking{}, \papillon{}, and \hover{} are presented below.
Example datapoints for each task are presented in \autoref{sec:experiments_lms_and_datasets}.
Code snippets assume respective DSPy imports are available.

\subsection{\bankingtext{}}
A DSPy program for \banking{} is shared in \autoref{snippetpython:banking_program}.
\begin{snippetpython}[caption={DSPy program for Banking77.}, label={snippetpython:banking_program}]
LABELS = [ ... ]  # List of all output labels
Banking77 = dspy.ChainOfThought(f"text -> label: Literal{LABELS}")
\end{snippetpython}

\subsection{\papillontext{}}
A DSPy program for \papillon{} is shared in \autoref{snippetpython:papillon_program}.
\begin{snippetpython}[caption={DSPy program for Papillon.}, label={snippetpython:papillon_program}]
class CraftRedactedRequest(dspy.Signature):
    """
    Given a private user query, create a privacy-preserving request for a powerful external LLM.
    The LLM may assist without learning private information about the user.
    """

    user_query = dspy.InputField()
    llm_request = dspy.OutputField()


class RespondToQuery(dspy.Signature):
    """
    Respond to a user query.
    For inspiration, we found a potentially related request to a powerful external LLM and its response.
    """

    related_llm_request = dspy.InputField()
    related_llm_response = dspy.InputField(desc="information from a powerful LLM responding to a related request")
    user_query = dspy.InputField(desc="the user's request you need to fulfill")
    response = dspy.OutputField(desc="your final response to the user's request")


class PAPILLON(dspy.Module):
    def __init__(self, untrusted_model):
        self.craft_redacted_request = dspy.ChainOfThought(CraftRedactedRequest)
        self.respond_to_query = dspy.Predict(RespondToQuery)
        self.untrusted_model = untrusted_model

    def forward(self, user_query):
        llm_request = self.craft_redacted_request(user_query=user_query).llm_request
        llm_response = self.untrusted_model(llm_request)[0]
        response = self.respond_to_query(
            related_llm_request=llm_request, related_llm_response=llm_response, user_query=user_query
        ).response

        return dspy.Prediction(llm_request=llm_request, llm_response=llm_response, response=response)
\end{snippetpython}

\subsection{\hovertext{}}
A DSPy program for \hover{} is shared in \autoref{snippetpython:hover_program}.
\begin{snippetpython}[caption={DSPy program for HoVer.}, label={snippetpython:hover_program}]
# Assume that a function called deduplicate is defined

class GenerateThreeQueries(dspy.Signature):
    """
    Given a claim and some key facts, generate up to 3 followup queries to find the next most essential clue towards verifying or refuting the claim. If you think fewer queries are sufficient, generate None for the search query outputs you don't need. The goal ultimately is to find all documents implicated by the claim.
    """
    claim = dspy.InputField()
    key_facts = dspy.InputField()
    search_query1 = dspy.OutputField()
    search_query2 = dspy.OutputField()
    search_query3 = dspy.OutputField()


class AppendNotes(dspy.Signature):
    """
    Given a claim, some key facts, and new search results, identify any new learnings from the new search results, which will extend the key facts known so far about whether the claim is true or false. The goal is to ultimately collect all facts that would help us find all documents implicated by the claim.
    """
    claim = dspy.InputField()
    key_facts = dspy.InputField()
    new_search_results = dspy.InputField()
    new_key_facts = dspy.OutputField()


class Hover(dspy.Module):
    def __init__(
            self,
            num_hops=4,
            k_per_search_query=10,
            k_per_search_query_last_hop=30,
            num_total_passages=100,
        ):
        # Value is fixed to simplify signature construction in presented snippet
        self.num_search_queries_per_hop = 3

        self.num_hops = num_hops
        self.k_per_search_query = k_per_search_query
        self.k_per_search_query_last_hop = k_per_search_query_last_hop
        self.num_total_passages = num_total_passages

        self.rm = dspy.ColBERTv2()
        self.generate_query = dspy.ChainOfThought(GenerateThreeQueries)
        self.append_notes = dspy.ChainOfThought(AppendNotes)

    def forward(self, claim: str) -> list[str]:
        key_facts = []
        committed_docs = []

        for hop_ind in range(self.num_hops):
            is_last_hop = hop_ind == self.num_hops - 1
            is_first_hop = hop_ind == 0
            hop_k = self.k_per_search_query_last_hop if is_last_hop else self.k_per_search_query
            num_docs_to_keep = (self.num_total_passages - len(committed_docs)) if is_last_hop else self.k_per_search_query

            if is_first_hop:
                search_queries = [claim]
            else:
                pred = self.generate_query(claim=claim, key_facts=key_facts)
                search_queries = [pred.search_query1, pred.search_query2, pred.search_query3]
            search_queries = deduplicate(search_queries)

            search_results = [r for q in search_queries for r in search_raw(q, k=hop_k, rm=self.rm)]
            search_results = sorted(search_results, key=lambda r: r["score"], reverse=True)

            unique_docs = []
            for result in search_results:
                if result["long_text"] not in unique_docs:
                    unique_docs.append(result["long_text"])
            unique_docs = unique_docs[:num_docs_to_keep]
            committed_docs.extend(unique_docs)

            if not is_last_hop:        
                pred = self.append_notes(claim=claim, key_facts=key_facts, new_search_results=unique_docs)
                key_facts.append(pred.new_key_facts)

        return dspy.Prediction(key_facts=key_facts, retrieved_docs=committed_docs)
\end{snippetpython}

\section{Example DSPy traces demonstrating disjoint module contexts}
\label{app:example_trace}

The following illustrates how DSPy executes a multi-module program where each module receives its own locally constructed prompt and thus operates over a disjoint context.
Note that for brevity we show a full trace from a simplified version of the \hover{} program shown in \autoref{app:task_details}.

\begin{chatmessage}[caption={\hover{} hop 1 (\text{key-fact extraction} module): system prompt.}, label={message:hover_hop1_system}]
Your input fields are:

1. `claim` (str):
2. `key_facts` (str):
3. `new_search_results` (str):

Your output fields are:
4. `reasoning` (str):
5. `new_key_facts` (str):

All interactions will follow this structure:

[[ ## claim ## ]]
{claim}

[[ ## key_facts ## ]]
{key_facts}

[[ ## new_search_results ## ]]
{new_search_results}

[[ ## reasoning ## ]]
{reasoning}

[[ ## new_key_facts ## ]]
{new_key_facts}

[[ ## completed ## ]]

Your objective:
Given a claim, some key facts, and new search results, identify any new learnings required to verify or refute the claim.
\end{chatmessage}

\begin{chatmessage}[caption={\hover{} hop 1 (\text{key-fact extraction} module): user prompt.}, label={message:hover_hop1_user}]
[[ ## claim ## ]]
Bruce Geller died in 1978. He developed the American television detective show that Gail Fisher plays the role of Peggy Fair in.

[[ ## key_facts ## ]]
N/A

[[ ## new_search_results ## ]]
[1] "Gail Fisher ... played Peggy Fair on the detective series Mannix ..."
[2] "Mannix ... developed by executive producer Bruce Geller ..."
[3] "Bruce Bernard Geller (1930-1978) was an American producer ..."
[4] "List of Mannix episodes ... developed by Bruce Geller ..."

Respond with [[ ## reasoning ## ]], [[ ## new_key_facts ## ]], and [[ ## completed ## ]].
\end{chatmessage}

\begin{chatmessage}[caption={\hover{} hop 1 (\text{key-fact extraction} module): assistant output.}, label={message:hover_hop1_assistant}]
[[ ## reasoning ## ]]
The search results confirm Bruce Geller's death date and his role in developing Mannix. They also confirm Gail Fisher's role as Peggy Fair. These facts clarify the relationships relevant to the claim.

[[ ## new_key_facts ## ]]
1. Bruce Geller (1930-1978) developed the detective show Mannix.
2. Gail Fisher (1935-2000) played Peggy Fair on Mannix (1968-1975).

[[ ## completed ## ]]
\end{chatmessage}

\begin{chatmessage}[caption={\hover{} hop 2 (\text{follow-up search} module): system prompt.}, label={message:hover_hop2_system}]
Your input fields are:

1. `claim` (str):
2. `key_facts` (str):

Your output fields are:
3. `reasoning` (str):
4. `followup_search_query` (str):

All interactions follow this structure:

[[ ## claim ## ]]
{claim}

[[ ## key_facts ## ]]
{key_facts}

[[ ## reasoning ## ]]
{reasoning}

[[ ## followup_search_query ## ]]
{followup_search_query}

[[ ## completed ## ]]

Your objective:
Given a claim and extracted key facts, generate a followup search query to find the next most essential clue for verification.
\end{chatmessage}

\begin{chatmessage}[caption={\hover{} hop 2 (\text{follow-up search} module): user prompt.}, label={message:hover_hop2_user}]
[[ ## claim ## ]]
Bruce Geller died in 1978. He developed the American television detective show that Gail Fisher plays the role of Peggy Fair in.

[[ ## key_facts ## ]]
<<<
1. Bruce Geller (1930-1978) developed Mannix.
2. Gail Fisher (1935-2000) played Peggy Fair on Mannix.
>>>

Respond with [[ ## reasoning ## ]], [[ ## followup_search_query ## ]], and [[ ## completed ## ]].
\end{chatmessage}

\begin{chatmessage}[caption={\hover{} hop 2 (\text{follow-up search} module): assistant output.}, label={message:hover_hop2_assistant}]
[[ ## reasoning ## ]]
The key facts give the dates and roles of Geller and Fisher. The next step is to verify the exact nature of Geller's contribution to developing Mannix.

[[ ## followup_search_query ## ]]
"Bruce Geller developed Mannix" "Mannix Gail Fisher Peggy Fair"

[[ ## completed ## ]]
\end{chatmessage}

\section{Asset information}
\label{app:assets}

The license information for the models and datasets we used is documented below.
All models and datasets are accessed via \href{https://huggingface.co/}{HuggingFace}.

\textbf{\qwenthreeshort{}} is released under the \apachetwolicense{}, accessed via the HuggingFace model identifier \qwenthree{}.

\textbf{\llamathreepointoneshort{}} is released under the \llamathreecommunitylicense{}, accessed via the HuggingFace model identifier \llamathreepointone{}.

\textbf{\banking{}} is released under the \ccbyfour{} license.

\textbf{\hover{}} is released under the \ccbyfour{} license.

\textbf{\papillon{}} is released under the \mitlicense{}.

\end{document}